\pdfoutput=1
\documentclass[11pt,a4paper,UTF8]{article}
\usepackage{CJKutf8}

\usepackage[]{naacl2021}
\usepackage[T1]{fontenc}
\usepackage[utf8]{inputenc}
\usepackage{times}
\usepackage{latexsym}
\usepackage{subfigure} 

\usepackage{microtype}
\usepackage{multirow}
\usepackage{graphics}
\usepackage{graphicx}
\usepackage{booktabs}
\usepackage{amsthm,amsmath,amssymb}
\usepackage{mathrsfs}

\newcommand{\textttf}[1]{{\small{\texttt{#1}}}}

\title{Lattice-BERT: Leveraging Multi-Granularity Representations in Chinese Pre-trained Language Models }

\author{
    Yuxuan Lai$^{1,2,}$\thanks{\;\;Work done during an internship at Alibaba DAMO Academy.}~~, 
    Yijia Liu$^{3}$, 
    Yansong Feng$^{1,2,}$\thanks{\;\;Corresponding author.}~~,  
    Songfang Huang$^{3}$ \and
    Dongyan Zhao$^{1,2}$ \\
    $^1$Wangxuan Institute of Computer Technology, Peking University, China\\
    $^2$The MOE Key Laboratory of Computational Linguistics, Peking University, China\\
    $^3$Alibaba Group \\
    {\tt \{erutan, fengyansong, zhaody\}@pku.edu.cn} \\
    {\tt \{yanshan.lyj, songfang.hsf\}@alibaba-inc.com}\\
}

\date{}

\begin{document}
\begin{CJK}{UTF8}{gbsn}
\maketitle
\begin{abstract}
Chinese pre-trained language models usually process text as a sequence of characters, while ignoring more coarse granularity, e.g., words. 
In this work, we propose a novel pre-training paradigm for Chinese --- Lattice-BERT, which explicitly incorporates word representations along with characters, thus can model a sentence in a multi-granularity manner. 
Specifically, we construct a lattice graph from the characters and words in a sentence and feed all these text units into transformers. 
We design a lattice position attention mechanism to exploit the lattice structures in self-attention layers.
We further propose a masked segment prediction task to push the model to learn from rich but redundant information inherent in lattices, while avoiding learning unexpected tricks.
Experiments on 11 Chinese natural language understanding tasks show that our model can bring an average increase of 1.5\% under the 12-layer setting, which achieves new state-of-the-art among \textit{base}-size models on the CLUE benchmarks. 
Further analysis shows that Lattice-BERT can harness the lattice structures, and the improvement comes from the exploration of redundant information and multi-granularity representations.\footnote{Our code is available at \url{https://github.com/alibaba/AliceMind/tree/main/LatticeBERT}.}
\end{abstract}

\section{Introduction}

Pre-trained Language Models (PLMs) have achieved promising results in many Chinese Natural Language Understanding (NLU) tasks \cite{bert-wwm,kbert,ernie2}.
These models take a sequence of fine-grained units --- Chinese characters --- as the input, following the English PLMs' practice \cite[BERT]{bert}.

However, the meanings of many Chinese words cannot be fully understood through direct compositions of their characters' meanings.
For example, 老板/\textit{boss} does not mean 老/\textit{elder} 板/\textit{board}.\footnote{For clarity, we use 中文/\textit{English translation} to represent an example in Chinese with its \textit{translation} followed by.}
The importance of word-level inputs in Chinese has been addressed in different tasks, including relation classification
\cite{li2019chinese}, short text matching \cite{lai2019lattice,chen-etal-2020-neural,Lyu2021LETLK}, trigger detection \cite{lin2018nugget}, and named entity recognition \cite{latticeLSTMner,gui2019lexicon,flat}.
The coarse-grained inputs benefit these tasks by introducing word-level semantics with multi-granularity representations,
which is potentially complementary in character-level Chinese PLMs.

\begin{figure}[t]
    \centering
    \includegraphics[width=0.97\columnwidth]{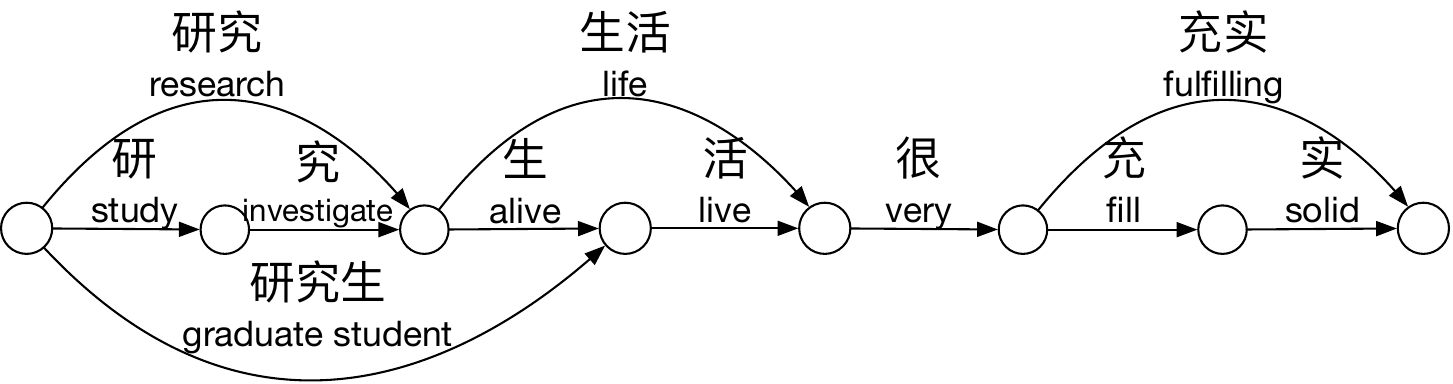}
\caption{ An illustration of the word lattice for sentence 研究生活很充实/\textit{Research Life is very fulfilling}. 
\label{fig:lattice}}
\end{figure}

\begin{figure*}[t]
    \centering
    \includegraphics[width=0.97\textwidth]{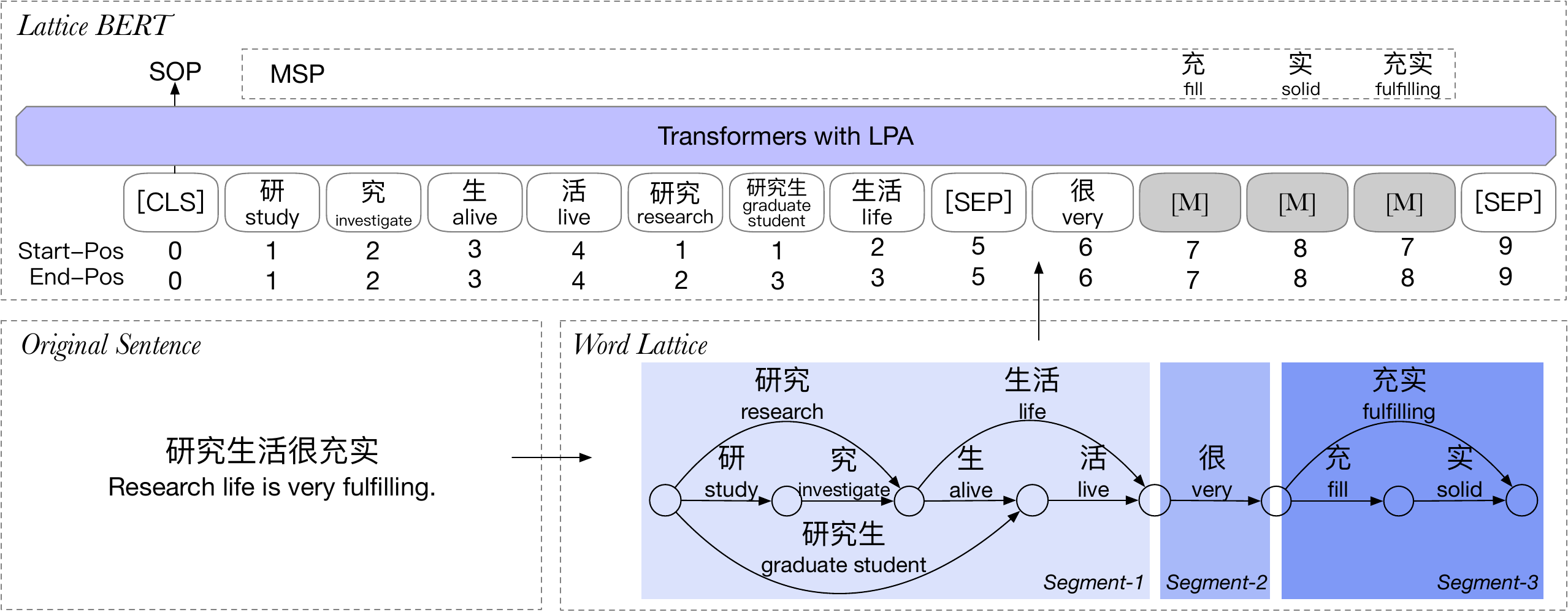}
\caption{An illustration of our pre-training framework, Lattice-BERT.
\label{fig:lattice-bert}
}
\end{figure*}

In this work, we discuss how to pre-train a Chinese PLM over a \textit{word lattice} structure to exploit multi-granularity inputs.
We argue that by incorporating the coarse-grained units into PLM, 
models could learn to utilize the multi-granularity information for downstream tasks.
Specifically, we organize characters and words in sentences as word lattices (see Figure~\ref{fig:lattice}), which enable the models to explore the words from all possible word segmentation results.

However, it is not straightforward to learn a BERT-like PLM over the word lattices.
The major challenges are two-folded.
Firstly, BERT's original input is a sequence of characters ordered by their positions, making it difficult to consume the word lattices and preserve the positional relationship between multi-granularity units.
Secondly, the conventional masked language modeling (MLM) task may make the word-lattice based PLMs learn unexpected tricks. The reason is that such a word lattice naturally introduces redundancy, that is, one character can be contained in multiple text units.
In MLM, models may refer to the other text units overlapping with the randomly masked one instead of the real context, which brings information leakages.

To address these challenges, we propose a Lattice-based Bidirectional Encoder Representation from Transformers (Lattice-BERT).
Specifically, we design a \textit{lattice position attention} (LPA) to help the transformers directly exploit positional relationship and distances between text units in lattices.
Moreover, we propose a \textit{masked segment prediction} (MSP) task to avoid the potential leakage between overlapping text units in language modeling.
With LPA and MSP, the Lattice-BERT could 
harness the multi-granularity structures in lattices, thus, 
directly utilize the lattice structures to aggregate the coarse-grained word information to benefit various downstream tasks.

We evaluate our model on 11 Chinese NLU tasks in various paradigms, including the CLUE benchmarks \cite{CLUEbenchmark} as well as two sequence labeling tasks.
Compared with the baseline that only takes characters as inputs, Lattice-BERT bring an average increase of 1.5\% and 2.0\% under the settings of 12 and 6 layers, respectively. 
The 12-layer Lattice-BERT model beats all other \textit{base}-size models on CLUE benchmarks.\footnote{\url{https://www.cluebenchmarks.com/rank.html}, until Oct. 31st, 2020.}
Morever, we show that Lattice-BERT can harness the multi-granularity inputs and utilize word-level semantics to outperform vanilla fine-grained PLMs.

Our contributions can be summarized as
1) We propose Lattice-BERT to leverage multi-granularity representations from word lattices in Chinese PLMs.
2) We design \textit{lattice position attention} and \textit{masked segment prediction} to facilitate Chinese PLMs to exploit the lattice structures.
3) Lattice-BERT brings remarkable improvements on 11 Chinese tasks and achieves new state of the arts among \textit{base}-size models at the CLUE benchmarks.

\section{Lattice-BERT}

This section, we detail the implementation of Lattice-BERT, and its overall framework is presented in Figure~\ref{fig:lattice-bert}.

\subsection{Preliminary: BERT}
 
BERT \cite[Bidirectional Encoder Representations from Transformers]{bert} is a pre-trained language model comprising a stack of multi-head self-attention layers and fully connected layers.
For each head in the $l_{th}$ multi-head self-attention layer, the output matrix $\mathbf{H}^{\mathrm{out}, l} =\left\{\mathbf{h}_1^{\mathrm{out}, l}, \mathbf{h}_2^{\mathrm{out}, l} , \dots, \mathbf{h}_{n}^{\mathrm{out}, l} \right\} \in \mathbb{R}^{n \times d_k} $ satisfies:
$$
\mathbf{h}_i^{\mathrm{out}, l} =
\sum_{j=1}^{n}{\left(\frac{\exp{\alpha_{ij}^l}}{\sum_{j'}{\exp{\alpha_{ij'}^l}}} \mathbf{h}_j^{\mathrm{in},l} \mathbf{W}^{v, l} \right)} 
$$
\begin{equation}
\alpha_{ij}^l=\frac{1}{\sqrt{2d_k}}\left(\mathbf{h}_i^{\mathrm{in},l}\mathbf{W}^{q, l}\right){\left(\mathbf{h}_j^{\mathrm{in},l}\mathbf{W}^{k, l}\right)}^T \label{eq1}
\end{equation}
where $\mathbf{H}^{\mathrm{in}, l} =\left\{\mathbf{h}_1^{\mathrm{in}, l}, \mathbf{h}_2^{\mathrm{in}, l} , \dots, \mathbf{h}_n^{\mathrm{in}, l} \right\} \in \mathbb{R}^{n \times d_h}$ is the input matrix, and $\mathbf{W}^{q, l}, \mathbf{W}^{k, l}, \mathbf{W}^{v, l} \in \mathbb{R}^{d_h \times d_k}$ are learnable parameters. $n$ and $d_h$ are sequence length and hidden size, and the attention size $d_k$ = $d_h / n_h$, where $n_h$ is the number of attention heads. 

To capture the sequential features in languages, previous PLMs adopt position embedding in either input representations \cite{bert,albert} or attention weights \cite{xlnet,wei2019nezha,ke2020rethinking}.
For the input-level position embedding, the inputs of the first layer are $\mathbf{\widetilde{h}}_i^{\mathrm{in},0} = \mathbf{{h}}_i^{\mathrm{in},0} + \mathbf{P}_i$, where $\mathbf{P}_i$ is the embedding of the $i_{th}$ position.
The other works incorporate position information in attention weights, i.e., $\widetilde{\alpha}_{ij}^l={\alpha}_{ij}^l + f\left(i, j\right)$, where $f$ is a function of the position pair $\left(i, j\right)$.

The BERT model is pre-trained on an unlabeled corpus with reconstruction losses, i.e., Masked Language Modeling (MLM) and Next Sentence Prediction (NSP), and then fine-tuned on downstream tasks to solve specific NLU tasks.
Readers could refer to \citet{bert} for details.

\subsection{Multi-granularity Inputs: Word Lattices}
We adopt a word lattice to consume all possible segmentation results of a sentence in one PLM.
Each segmentation can be a mixture of characters and words.
As shown in Figure~\ref{fig:lattice}, a word lattice is a directed acyclic graph, where the nodes are positions in the original sentences, and each directed edge represents a character or a plausible word.
Word lattices incorporate all words and characters so that models could explicitly exploit the inputs of both granularities, despite some of the words are redundant.
In the rest of this work, we use \textit{\textbf{lattice tokens}} to refer to text units, including the characters and words, contained in lattice graphs.

As shown in Figure~\ref{fig:lattice-bert}, we list the lattice tokens in a line and consume these tokens to transformers straightforwardly.
However, the challenges of learning PLMs as BERT over the lattice-like inputs include:
1) encoding the lattice tokens while preserving lattice structures; 
2) avoiding potential leakage brought by redundant information. 

\subsection{Interaction: Lattice Position Attention}

Since the original BERT is designed for sequence modeling, it is not straightforward for BERT to consume a lattice graph.
The word lattices encode not only the character sequences but also nested and overlapping words from different segmentations.
To accurately incorporate positional information from lattice graphs into the interactions between tokens, we extend the attention-level position embedding and propose \textit{lattice position attention}. 

The \textit{lattice position attention} aggregates the attention score of token representations, $\alpha_{ij}$ in Eq.~\ref{eq1}, with three position related attention terms,
encoding the absolute positions, the distance, and the positional relationship, which can be formulated as:
\begin{equation}
\widetilde{\alpha}_{ij} = {\alpha}_{ij} + \mathrm{att}_{ij} + b_{ij} + r_{ij} 
\label{eq2}
\end{equation}

The $\mathrm{att}_{ij}$ in Eq.~\ref{eq2} is the attention weight between the absolute positions:
$$
\mathrm{att}_{ij} = \frac{1}{\sqrt{2 d_k}}
\left(\left[\mathbf{P}^S_{s_i};\mathbf{P}^E_{e_i}\right] \mathbf{W}^{q}\right)
\left(\left[\mathbf{P}^S_{s_j};\mathbf{P}^E_{e_j}\right] \mathbf{W}^{k}\right)^T
$$
$\left[ \cdot ; \cdot  \right]$ means the concatenation of vectors. $\mathbf{W}^{q}, \mathbf{W}^{k} \in \mathbb{R}^{2d_e \times d_k}$ are learnable parameters, $d_e$ and $d_k$ are embedding size and attention size.
$s_i, e_i$ are positions of start and end characters of the $i_{th}$ token. 
Taking the word  研究/\textit{research} in Figure~\ref{fig:lattice} as an example, it starts at the first character and ends at the second one, thus, its $s_i$ and $e_i$ are 1 and 2, respectively.
$\mathbf{P}^S$ and $\mathbf{P}^E$ are learnable position embedding matrices. 
$\mathbf{P}^S_t, \mathbf{P}^E_t \in \mathbb{R}^{d_e} $ is the $t_{th}$ embedding vector of $\mathbf{P}^S$ or $\mathbf{P}^E$. 
The $\mathrm{att}_{ij}$ exploit the prior of attention weight between the start and end positions of the token pairs.

The $b_{ij}$ in Eq.~\ref{eq2} is the attention term for the distance between the $i_{th}$ and $j_{th}$ tokens, which consists of four scaling terms considering the combinations of the start and end positions:
$$
b_{ij} = 
{b}^{ss}_{{s_j}-{s_i}} + 
{b}^{se}_{{s_j}-{e_i}} + 
{b}^{es}_{{e_j}-{s_i}} + 
{b}^{ee}_{{e_j}-{e_i}}
$$
$b^{ss}_t$ reflects the attention weight brought by the relative distance $t$ between the start positions of two tokens.
The other terms, i.e., $b^{se}_t$, $b^{es}_t$, and $b^{ee}_t$, have similar meanings.
In practice, the distance \textit{t} is clipped into $\left[-128, 128\right]$.

\begin{figure}[t!]
    \centering
    \includegraphics[width=0.48\textwidth]{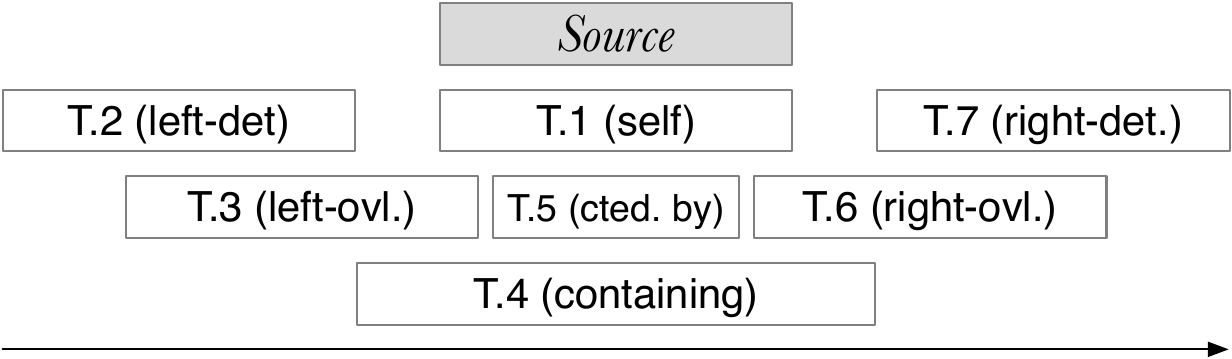}
\caption{An illustration of the positional relations. Each rectangle represents a lattice token, corresponding to a character span in the original sentence. \textit{T.1} $\sim$ \textit{T.7} are the target tokens with the seven different position relation to the \textit{Source} token.
\label{fig:pos-rel}
}
\end{figure}

$r_{ij}$ in Eq.~\ref{eq2} is a scaling term represents the positional relation between the $i_{th}$ and $j_{th}$ tokens. 
We consider seven relations, including (1) self, (2) left and detached, (3) left and overlapped, (4) containing, (5) contained by, (6) right and overlapped, (7) right and detached. 
Figure~\ref{fig:pos-rel} shows an illustration of these 7 relations. Formally, 
for the $i_{th}$ and $j_{th}$ tokens, they are overlapped means $s_i \leq s_j < e_i \leq e_j$ or $s_j \leq s_i < e_j \leq e_i$, and if $e_i < s_j$ or $e_j < s_i$, they are detached. 
If $s_i \leq s_j \leq e_j \leq e_i$ and $i \neq j$, the $i_{th}$ token contains the $j_{th}$ token and the $j_{th}$ token is contained by the $i_{th}$ token.
Intuitively, only two detached tokens can be concurrent in one Chinese word segmentation result. Moreover, the containing relation reflects a sort of lexical hierarchy in the lattices.
We think $r_{ij}$ can explicitly model the positional relations between tokens in lattice graphs. 

We argue that the attention scores for distances and token relations capture different aspects of the multi-granularity structures in lattice graphs, thus, meeting the needs of various downstream tasks, such as distance for coreference resolution and positional relation for named entity recognition. 
With the information of absolute positions, distances, and positional relations, PLMs could accurately exploit the lattice structures in attention layers.

The \textit{lattice position attention} weights are shared over all layers. 
$b_{ij}$, $r_{ij}$, $\mathbf{W}^q$, and $\mathbf{W}^k$ are diverse in different attention heads to capture diverse attention patterns.
We follow \citet{ke2020rethinking} to reset the positional attention scores related to \textttf{[CLS]} tokens, which is the special token as prefix of the input sequences to capture the overall semantics.

\subsection{Pre-training Tasks: Masked Segment Prediction}

Vanilla BERT is trained to predict the randomly masked tokens in the sentences, i.e., the masked language modeling (MLM).
For the case of consuming multi-granularity inputs, the input tokens are redundant
which means a character can occur in its character forms and multiple words it belongs to.
Directly adopting the randomly masking strategy 
may simplify the prediction problem in our case because the masked token can be easily guessed
via peeking the unmasked tokens overlapping with the masked one.
Taking the word 研究/\textit{research} in Figure~\ref{fig:lattice-bert} as an example,
supposing the masked input is \texttt{[M]}/究/研究,
the model will consult 研究 rather than the context to predict the masked token, 研.

We investigate this problem and find that the tokens within a minimal \textit{segment} of the lattice provide strong clue for the prediction of other tokens.
A \textit{\textbf{segment}} is a connected subgraph of a lattice where no token exists outside the subgraph that overlaps with any token inside the subgraph.
To identify these minimal segments, we enumerate the character-level tokens in sentence order, checking if all the word-level tokens which contain this character end at this character.
If so, all the tokens containing previous and current characters are considered as a segment, and the next segment starts from the next character, see the example in Figure~\ref{fig:lattice-bert}. 
After the segment detection, we propose a \textit{masked segment prediction} (MSP) task as a replacement of the MLM in the original BERT.
In MSP, we mask out all the tokens in a segment and predict all these tokens (see Figure~\ref{fig:lattice-bert}) to avoid the potential leakage. 

In addition to MSP, we also pre-train our models with the sentence order prediction (SOP) task in \citet{albert}, where the model predicts whether two consecutive sentences are swapped in inputs.

\subsection{Downstream tasks with Lattice-BERT}

We explore four kinds of downstream tasks, i.e., sentence/sentence-pair classification, multiple choices, sequence labeling, and span selection machine reading comprehension (MRC).
For the sentence/sentence-pair classification, both vanilla and Lattice-BERT classify input instances base on logistic regressions over the representation of \textttf{[CLS]} tokens in the last layer.
The circumstances are similar in multiple choice tasks, where softmax regressions are conducted over the representations of \textttf{[CLS]} tokens to choose the best options.
However, for the span selection MRC, and the sequence labeling tasks like named entity recognition (NER), models need to perform token-wise classification.
Vanilla BERT predicts labels for the input characters, but lattice-BERT has additional words.
In Lattice-BERT, we extract the character chains (word pieces for numbers and English words) from lattices for training and prediction for a fair comparison with vanilla BERT.
Pilot studies show that this strategy performs comparably with the more complex strategies, which supervise the labels over words and obtain a character's label via ensembles of all tokens containing that character.

\subsection{Implementation}

\paragraph{Lattice Construction.}
We construct the word lattices based on a vocabulary consisting of 102K high-frequency open domain words.
All the sub-strings of the input sequence that appear in the vocabulary are considered lattices tokens of the input.
With Aho-Corasick automaton \cite{aho1975efficient}, this construction procedure can complete in linear time to the size of the corpus and the vocabulary.\footnote{Formally, the time complexity is $O((N+M) \times L)$.
$N$ is the corpus size, $M$ is the vocabulary size, and $L$ is the average length of the words in characters, which is a small constant.} 
To deal with English words and numbers where the substrings are meaningless, we use the character sequences for those out-of-vocabulary non-Chinese inputs and remain the in-vocabulary words and word pieces.

We construct word lattices using all possible words according to a vocabulary instead of more sophisticated lattice construction strategies. 
Previous research efforts~\cite{lai2019lattice,chen-etal-2020-neural,bertmwa} on lattice construction suggests that using all possible words usually yields better performance.
We think an overly-designed lattice construction method may bias our model on certain types of text, and would probably harm the generalization. 
So, in our case, we let the model learn by itself to filter the noise introduced by using \textit{all possible words} during pre-training on a large-scale corpus.

\paragraph{Pre-training Details.}
To compare with previous pre-training works, we implement the \textit{base}-size models, which contains 12 layers, 768-dimensional of hidden size, and 12 attention heads. 
To demonstrate how lattice gains in shallower architectures and provide lightweight baselines, we also conduct the \textit{lite}-size models with 6 layers, 8 attention heads, and the hidden size of 512.

To avoid the large vocabulary introducing too many parameters in embedding matrix, we adopt the embedding decomposition trick following \citet[ALBERT]{albert}. 
Consequently, the parameters of Lattice-BERT is 100M in base-size, only 11\% more than its character-level counterpart (90M), and smaller than the RoBERTa-base \cite{roberta} (102M) and AMBERT \cite{zhang2020ambert} (176M).
The modeling of positional relation and distances in \textit{lattice position attention} introduces only 12K parameters.

A collection of Chinese text, including Chinese Wikipedia, Zhihu, and web news, is used in our BERT models' pre-training stage. 
The total number of characters in our unlabeled data is 18.3G.
We follow \citet{roberta} and train the PLMs with a large batch size of 8K instances for 100K steps.
The hyper-parameters and details are given in Appendix~\ref{app:imp}.

\section{Experiments}

We present the details of the Lattice-BERT fine-tuning results on 11 Chinese NLU tasks.
Answering the following questions: 
(1) Whether the Lattice-BERT performs better than mono-granularity PLMs and other multi-granularity PLMs?
(2) How the proposed \textit{lattice position attention} and \textit{masked segment prediction} contribute to the downstream tasks? 
(3) How Lattice-BERT outperforms the original character-level PLMs?

\subsection{Tasks}
We test our models on 11 Chinese NLU tasks, including the text classification and Machine Reading Comprehension (MRC) tasks in the Chinese Language Understanding Evaluation benchmark \cite[CLUE]{CLUEbenchmark}, and two additional tasks to probe the effectiveness in sequence labeling.

\textbf{CLUE text classification}:
natural language inference \textbf{CMNLI}, 
long text classification IFLYTEK (\textbf{IFLY.}), short text classification \textbf{TNEWS},  
semantic similarity \textbf{AFQMC}, coreference resolution (CoRE) CLUEWSC 2020 (\textbf{WSC.}), and key word recognition (KwRE) \textbf{CSL}.

\textbf{CLUE MRC}: 
Span selection based MRC CMRC 2018 (\textbf{CMRC}), multiple choice questions \textbf{C$^3$}, and idiom cloze \textbf{ChID}.

\textbf{Sequence Labeling}:
Chinese word segmentation (CWS) \textbf{MSR} dataset from SIGHAN2005 \cite{emerson2005second}, and 
named entity recognition (NER) \textbf{MSRA-NER} \cite{levow2006third}.

We probe our proposed Lattice-BERT model thoroughly with these various downstream tasks.
The statistics and hyper-parameters of each task are elaborated in Appendix~\ref{app:data}.
We tune learning rates on validation sets and report test results with the best developing performances for CLUE tasks.\footnote{We report accuracy / exact-match scores for CLUE tasks, and label-F1 / F1 for NER / CWS tasks. CLUE results see \url{https://www.cluebenchmarks.com/rank.html}.}
For MSR and MSRA-NER, we run the settings with the best learning rates five times and report the \textbf{\textit{average scores}} to ensure the reliability of results.

\subsection{Compared Systems}

\begin{table*}[h!]
\small
\centering
\setlength{\tabcolsep}{4.5pt}
\begin{tabular}{l cccccc|c|ccc|c|cc|c}
\toprule
\multirow{3}{*}{Task} & \multicolumn{7}{c|}{CLUE-Classification} & \multicolumn{4}{c|}{CLUE-MRC} & \multicolumn{2}{c|}{\scriptsize Seq. Labeling} & \multirow{3}{*}{\shortstack{\\avg.}}  \\ \cmidrule{2-14} 
 &
  NLI &
  \multicolumn{2}{c}{TC} &
  \multicolumn{1}{c}{SPM} &
  \multicolumn{1}{c}{CoRE} &
  \multicolumn{1}{c|}{KwRE} &
  \multirow{2}{*}{ \shortstack{\scriptsize avg.}} &
  \multicolumn{3}{c|}{MRC} &
  \multirow{2}{*}{ \shortstack{\scriptsize avg.}} &
  CWS &
  \multicolumn{1}{c|}{NER} &
   \\ 
  \cmidrule{2-7}
  \cmidrule{9-11}
  \cmidrule{13-14}
 & 
  \multicolumn{1}{c}{\scriptsize CMNLI} &
  \scriptsize TNEWS &
  \multicolumn{1}{c}{\scriptsize IFLY.} &
  \scriptsize AFQMC &
  \multicolumn{1}{c}{\scriptsize WSC.} &
  \multicolumn{1}{c|}{\scriptsize CSL} & &
  \scriptsize CMRC &
  \scriptsize ChID &
  \scriptsize C$^3$ & &
  MSR &
  \multicolumn{1}{c|}{\scriptsize MSRA}
   \\ \midrule
\multicolumn{15}{c}{\textit{base}-size settings} \\
\midrule
 RoBERTa &
  80.5 &
  67.6 &
  60.3 &
  74.0 &
  76.9 &
  84.7 &
  74.0 &
  \textbf{75.2} &
  83.6 &
  66.5 &
  75.1 &
  98.2 &
  96.8 &
  78.5 \\ 
NEZHA &
  81.1 &
  67.4 &
  59.5 &
  74.5 &
  - &
  83.7 &
  - &
  72.2 &
  84.4 &
  71.8 &
  76.1 &
  - &
  - &
  - \\  
BERT-word &
  80.0 &
  68.2 &
  60.0 &
  73.5 &
  75.5 &
  85.2 &
  73.7 &
  41.3 &
  80.9 &
  67.0 &
  63.1 &
  - &
  - &
  - \\
AMBERT &
  \textbf{81.9} &
  \textbf{68.6} &
  59.7 &
  73.9 &
  78.3 &
  \textbf{85.7} &
  74.7 &
  73.3 &
  \textbf{86.6} &
  69.6 &
  76.5 &
  - &
  - &
  - \\ \midrule
 BERT-Our &
  80.3 &
  67.7 &
  62.2 &
  74.0 &
  79.3 &
  81.6 &
  74.2 &
  72.7 &
  84.1 &
  68.6 &
  75.1 &
  98.4 &
  96.5 &
  78.7 \\ 
LBERT &
  81.1 &
  68.4 &
  \textbf{62.9} &
  \textbf{74.8} &
  \textbf{82.4} &
  84.0 &
  \textbf{75.6} &
  74.0 &
  \textbf{86.6} &
  \textbf{72.7} &
  \textbf{77.8} &
  \textbf{98.6} &
  \textbf{97.1} &
  \textbf{80.2}  \\ 
  \midrule
\multicolumn{15}{c}{\textit{lite}-size settings} \\
\midrule
 BERT-Our &
  77.9 &
  66.7 &
  60.7 &
  72.1 &
  62.4 &
  78.7 &
  69.7 &
  68.3 &
  78.7 &
  61.6 &
  69.5 &
  98.1 &
  95.5 &
  74.6  \\ 
 LBERT &
  79.1 &
  68.2 &
  61.9 &
  72.4 &
  70.0 &
  81.9 &
  72.3 &
  69.9 &
  81.3 &
  63.6 &
  71.6 &
  98.4 &
  96.2 &
  76.6  \\ 
\bottomrule
\end{tabular}
\caption{\label{Table-general}
The results on testing sets of 11 Chinese tasks. The \textbf{bold numbers} are the best scores in each column. 
}
\end{table*}

\textbf{RoBERTa} \cite{cui2020revisiting} is the Chinese version RoBERTa model \cite{roberta}, which adopts the whole word masking trick and external pre-training corpus, known as RoBERTa-wwm-ext.\footnote{\url{https://huggingface.co/hfl/chinese-roberta-wwm-ext}} \\
\textbf{NEZHA} \cite{wei2019nezha} is one of the best Chinese PLMs with a bag of tricks, which also explores attention-level position embedding. \\
\textbf{AMBERT} \cite{zhang2020ambert} is the state-of-the-art multi-granularity Chinese PLM, with two separated encoders for words and characters. \\
\textbf{BERT-word} is a Chinese PLM baseline, taking words as single-granularity inputs. We obtain the results from \citet{zhang2020ambert} directly. \\
\textbf{BERT-our} is our implemented BERT model, with the same pre-training corpus, model structures, hyper-parameters, and training procedure with Lattice-BERT, but taking characters as inputs. We also adopt the whole word masking trick. \\
\textbf{LBERT} is our proposed Lattice-BERT model, with word lattices as inputs, equipping with \textit{lattice position attentions} and \textit{masked segment prediction}.

\subsection{Main Results}

In Table~\ref{Table-general}, 
we can see in text classification, MRC, and sequence labeling tasks, with both \textit{base} and \textit{lite} sizes, LBERT works better than our character-level baselines consistently. 
LBERT-\textit{base} outperforms all previous \textit{base}-size PLMs in average scores and obtain the best performances in 7 of the 11 tasks.

Comparing with the mono-granularity PLMs in \textit{base}-size, LBERT takes benefits from word-level information and outperforms its character-level counterpart, BERT-our, by 1.5\% averagely.
Meanwhile, LBERT performs better than the word-level model, BERT-word, remarkably on CLUE tasks. 
We think the lattice inputs incorporate coarse-grained semantics while avoiding segmentation errors by combining multiple segmentation results.
Therefore, with the multi-granularity treatments in word lattices, PLMs obtain better performances in downstream tasks than the mono-granularity settings.

Furthermore, LBERT outperforms the previous state-of-the-art (\textit{sota}) multi-granularity PLM, AMBERT \cite{zhang2020ambert}, by 0.9\% in text classification and 1.3\% in MRC, averagely. 
Different from modeling the characters and words separately, the graph representations of word lattices could enhance the interaction between multi-granularity tokens and utilize all possible segmentation results simultaneously.
As a result, LBERT achieves a new \textit{sota} among the \textit{base}-size models on the CLUE leaderboard as well as the sub-leaderboards for text classification and MRC tasks.\footnote{\url{https://www.cluebenchmarks.com/rank.html}, the \textit{sota} by the time of submission, Oct. 31st, 2020.}

With \textit{lite}-size settings, LBERT brings 2.0\% improvement over BERT-our on average, which is larger than the case in \textit{base}-size.
In CWS, TNEWS, and CSL, the \textit{lite}-size LBERT even outperforms the \textit{base}-size BERT-our.
With more coarse-grained inputs, the shallower architectures do not require complicated interactions to identify character combinations but utilizing word representations explicitly, 
thus, narrowing the gap with the deeper ones.

\subsection{Analysis}

\paragraph{Ablation Study.} We conduct ablation experiments to investigate the effectiveness of our proposed \textit{lattice position attention} (LPA) and \textit{masked segment prediction} (MSP) in downstream tasks.
To reduce the computational costs, we base our pre-training settings on \textit{lite}-size with the sequence length of 128 characters. 
We select one task from each of the task clusters. We use the entity-level F1 score for NER to highlight the influence on boundary prediction.
We report the \textit{\textbf{average scores}} over 5 runs and use the development sets for CLUE tasks.

\begin{table}[t]
\small
\centering
\setlength{\tabcolsep}{5pt}
\begin{tabular}{l cccc}
\toprule
& WSC. & NER.~(EF1) & CMRC & avg. \\\midrule
BERT-our & 
66.3{\tiny~(1.9)}&
92.8{\tiny~(0.2)} &
57.2{\tiny~(0.7)} & 
72.1 \\
LBERT & 
75.3{\tiny~(1.3)} & 
94.1{\tiny~(0.1)} &
64.5{\tiny~(0.5)} & 
78.0\\
\quad --Rel. &
75.7{\tiny~(1.2)} & 
93.7{\tiny~(0.2)} &
63.8{\tiny~(0.8)} & 
77.7\\
\quad --Dis. & 
73.8{\tiny~(0.7)} & 
93.8{\tiny~(0.2)} &
63.1{\tiny~(0.4)} & 
76.9\\
\quad --Dis. --Rel. & 
72.8{\tiny~(0.5)} & 
93.6{\tiny~(0.1)} &
61.7{\tiny~(0.4)} & 
76.0\\
\quad --MSP & 
72.2{\tiny~(1.0)} & 
93.9{\tiny~(0.1)} &
63.0{\tiny~(0.7)} & 
76.4\\
\bottomrule
\end{tabular}
\caption{\label{Table-ablation}
Ablation results in \textit{lite}-size settings with standard deviation in subscripts. 
EF1 is the entity-level F1 score.
-Dis. and -Rel. represent the ablation of the relative distances and positional relations in LPA, respectively.
The small numbers in brackets are standard deviation scores over five runs.
}
\end{table}

We can see in Table~\ref{Table-ablation} that the ablation of either module (--Dis.--Rel. \& --MSP) leads to a substantial drop in the average scores.
In particular, replacing MSP with vanilla MLM, the average score of --MSP drops by 1.6\%.
For the WSC. task, where long-range dependency is required to resolve the coreference, the gap is high up to 3.1\%.
We trace this drop into the pre-training procedure and observe the MLM accuracy for the --MSP setting on the development set is 88.3\%.
However, if we mask the tokens within the segment and avoid potential leakages, the accuracy drastically drops to 48.8\%, much lower than the performance of LBERT training with MSP (56.6\%).
This gap provides evidence that the MSP task prevents the PLMs from tricking the target by
peeking the overlapping text units in one segment, thus encourages the PLMs to characterize the long-range dependency.

For the LPA method, without the positional relation (--Rel.), the entity-level F1 score on NER decreases by 0.4\%, and the performance on CMRC decreases by 0.7\%. 
The performance drops are similar to the case without distance information (--Dis.).
Without either of them (--Dis. --Rel.), the gaps widen to 0.5\% and 2.8\%, respectively.
The boundary predictions in NER and CMRC are more sensitive to the local linguistic structures like nested words or overlapping ambiguity.
With the positional relation and distance charaterized in attention, LBERT could accurately model the interaction between the nested and overlapping tokens in different segmentation results.
Meanwhile, the accuracy of WSC. remarkably drops without distance information.
The performance drops by 7.5\% and 5.8\% when the number of characters between the pronouns and candidate phrases is larger than 30, or between 20 to 30, respectively. 
For the rest cases, the drop is only 0.4\%.
With explicitly modeling of distance, LBERT predicts the long-distance coreference relations more accurately.
Averagely, without the positional relation and distance modeling in LPA, the performance drops by 2.0\% on the three tasks, showing the importance of LPA in assisting the PLMs to exploit the multi-granularity structures in word lattices.

\paragraph{How LBERT Improves Fine-grained PLMs?}
We compare the prediction results of LBERT and the character-level BERT-our in \textit{base}-size on development sets to investigate how the LBERT outperforms the vanilla fine-grained PLMs. 
Intuitively, the word-level tokens in lattices provide coarse-grained semantics, which argument the character-level inputs.

We observe in TNEWS, the short text classification task, LBERT brings more improvement in the shorter instances, where the statements may be too short to provide enough context for predictions. 
By dividing the development set into five bins with equal size according to the sentence length, LBERT outperforms BERT-our by 2.3\% and 1.3\% in the shortest and second shortest bins, respectively, 
larger than the average gain on the rest instances (0.6\%).
We think the redundant tokens in word lattices provide rich context for the semantics of these short statements.
For example, for the short title 我们村的电影院/\textit{the cinema in our village}, with the redundant words, 电影/\textit{movie}, 影院/{cinema}, and 电影院/\textit{cinema}, introduced in the lattice, LBERT classifies the instance as entertainment news instead of news stories.

Another case is the CSL task, where the target is to predict whether the candidate words are keywords for a given paragraph. 
For those instances, where LBERT 
identifies more than two word-level tokens from each candidate word averagely,
which accounts for 47\% of the dataset, the performance gain is 3.0\%, significantly larger than the average improvement of the rest, 1.0\%. 
We think LBERT understands the key words from various aspects by exploiting the redundant expressions in lattices.
For example, from the keyword candidate 太阳能电池/\textit{solar battery}, the 太阳/{solar}, 太阳能/{solar energy}, 电池/{battery}, and 太阳能电池/\textit{solar battery} are lattice tokens.
With these word-level tokens, LBERT could match this candidate with the expressions in the paragraph like 阳极/\textit{positive electrode}, 光/\textit{light}, 电子/\textit{electron}, 离子/\textit{ion}, etc.

On the other side, for MSRA-NER, 
LBERT reduces the errors in identifying entities with nested structures.
Averagely, the number of error cases where the predicted entities are nested with the golden ones are reduced by 25\% in LBERT.
For example, the organization entity 解放巴勒斯坦运动/\textit{Palestine National Liberation Movement} is nested with the location entity 巴勒斯坦/\textit{Palestine} and ends with an indicator to organizations, 运动/\textit{movement}.
The character-level baseline model mistakenly recognizes the 巴勒斯坦/\textit{Palestine} and 动/\textit{move} as a location and an organization, separately.
While LBERT identifies this entity correctly after integrating the words, 解放/\textit{liberate}, 巴勒斯坦/\textit{Palestine}, and 运动/\textit{movement}.
With the pre-trained multi-granularity representations, LBERT fuse the contextual information from words and characters simultaneously, and detects the correct entity in success.

\begin{figure}[t]
    \centering
    \includegraphics[width=0.51\textwidth]{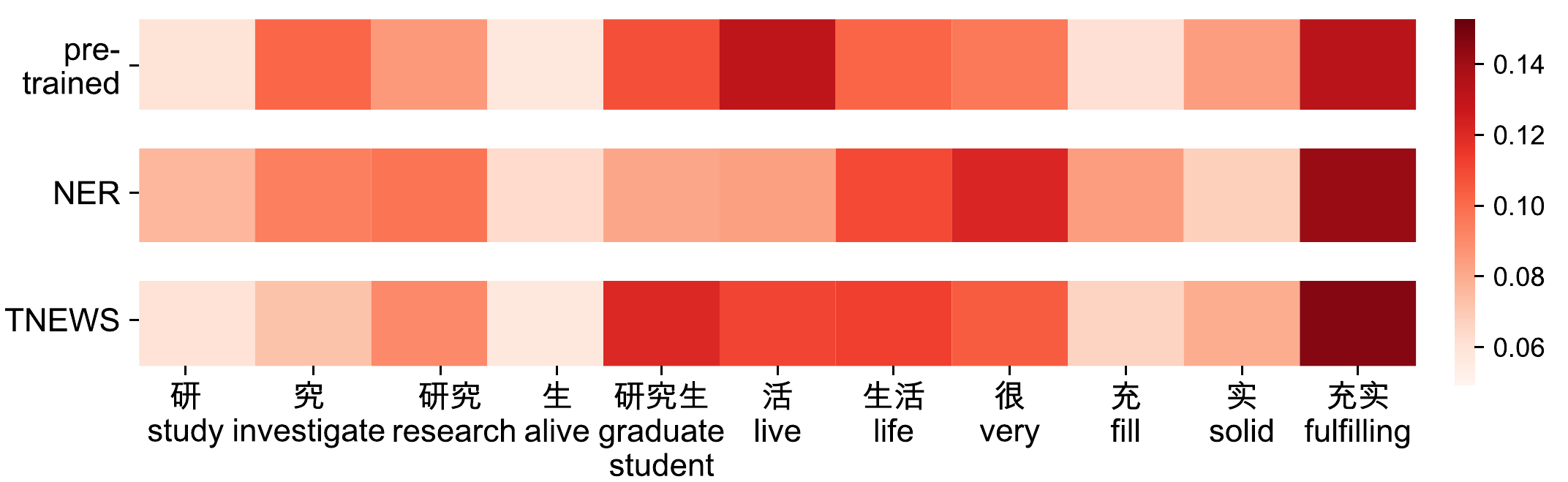}
\caption{Visualization of the attention scores of  研究生活很充实/\textit{Research life is very fulfilling}. 
The three lines from top to bottom are the attentions before fine-tuning and after fine-tuning with MSRA-NER and TNEWS tasks, respectively.
\label{fig:att-diff}
}
\end{figure}

\paragraph{How does LBERT harness multi-Granularity representations?}
LBERT consumes all the words and characters from input sequences simultaneously, but how does the model utilizes such multi-granularity representations during pre-training and downstream tasks?
To investigate this, we use the average attention scores that each lattice token receives among all layers and all heads to represent its importance.

As the example shown in Figure~\ref{fig:att-diff},
before fine-tuning, LBERT focuses on tokens including
活/\textit{live}, 充实/ \textit{fulfilling}, 研究/\textit{research},  研究生/\textit{graduate student}, 究/\textit{investigate}, etc.
Before fine-tuning on specific tasks, the model captures various aspects of the sentence.
After fine-tuning with MSRA-NER, the most focused words become 充实/\textit{fulfilling}, 很/\textit{very}, 生活/\textit{life}, and 研究/\textit{research}, i.e., the tokens from the golden segmentation result, ``研究|生活|很|充实'', which is intuitively beneficial for the NER tasks.
The attention score of the wrong segmented word, 研究生/\textit{graduate student}, drops remarkably. 

On the other hand, after fine-tuning with the news title classification task, TNEWS, LBERT tends to focus on 充实/\textit{fulfilling}, 研究生/\textit{graduate student}, 生活/\textit{life}, etc.
Although these tokens can not co-exist in one Chinese word segmentation result, LBERT can still utilize the redundant information from various plausible segmentations to identify the topics of inputs.
These results indicate that Lattice-BERT can well manage the lattice inputs by shifting the attention to different aspects among the multi-granularity representations according to specific downstream tasks.

\begin{table}[]
\setlength{\tabcolsep}{4pt}
\small
\centering
\begin{tabular}{p{7.5cm}}
\toprule
\textbf{Question}: What is the game with the theme song which is sung by Chen Yiting and is composed by Zeng Zhihao? \\ \midrule
\textbf{Document}: $\cdots$ He was recommended by Lu Shengfei, the No.7 music director of the Cape, as the composer of the theme song \textbf{\textit{The South of China}}$_{\left[\textrm{err}\right]}$. $\cdots$
In cooperation with singer Chen Yiting, Zeng Zhihao was responsible for the composition of the theme song Wish and the promotion song Youth Love for \textbf{\textit{The Legend of Sword and Fairy V}}$_{\left[\textrm{ans}\right]}$. \\
\bottomrule
\end{tabular}
\caption{ An example in CMRC. Given the \textbf{Question}, LBERT predicts the \textbf{\textit{corrected answer}}$_{\left[\textrm{ans}\right]}$ in the \textbf{Document}, while BERT-our predicts the \textbf{\textit{wrong answer}}$_{\left[\textrm{err}\right]}$.
\label{tab:case_study}}
\end{table}

\paragraph{Case Study.}
Table~\ref{tab:case_study} shows an example in CMRC, a span selection MRC task, where models choose a text span from the given document to answer the question.
In this case, the question asks for a \textit{game}, restricted by its \textit{theme song}.
BERT-our incorrectly outputs a theme song, \textit{The Song of China}, since there is no expression in the document explicitly related to \textit{game}.
However, LBERT find the correct answer, \textit{The Legend of Sword and Fairy V}. 
One possible reason is that \textit{The Legend of Sword and Fairy} is an entry in the vocabulary for lattice construction. 
LBERT may have learned this word as an entity for a famous video game from the context in pre-training by explicitly exploiting its representation as a whole.
With the coarse-grain text units in pre-training, LBERT directly encodes knowledge about these units to benefit the downstream tasks.

\paragraph{Computational Costs. }
For fair comparisons, we ensure LBERT and the character-level baselines (i.e., BERT-our) have the same training epochs when the training steps are equal following previous works \cite{zen,zhang2020ambert}.
Thus, comparing with BERT-our, 35\% more text units are introduced in the pre-training instances of LBERT, which introduces 48\% more computational resources comparing with BERT-our to process the additional word-level tokens (See Appendix~\ref{app:imp}). 
To illustrate the gains attribute to the incorporation of lattices instead of additional computations, we investigate the \textit{lite}-size BERT-our with longer input sequences in pre-training, which has the same computational costs as LBERT.
We find LBERT still outperforms BERT-our by 2.2\% averagely on CLUE classification tasks.
More details are elaborated in Appendix~\ref{app:exp}.

\section{Related Works}

Recently, several works utilize the lattice structures to explore multi-granularity information in Chinese NLU tasks.
\citet{buckman-neubig-2018-neural} incorporate lattice into recurrent neural networks based language modeling to capture marginal possibilities across all possible paths.
In NER, Lattice-LSTM \cite{latticeLSTMner}, graph neural networks \cite{gui2019lexicon}, and flat-lattice transformers \cite{flat} are adopted to incorporate words from lattice inputs.
\citet{lai2019lattice} adapt convolutional neural networks to lattice for matching based question answering.
\citet{chen-etal-2020-neural} adopt graph matching networks to perform multi-granularity interaction between lattices for text similarity.
These works are designed to explore word lattices in specific tasks. 
We explore the multi-granularity representations with word lattices in PLMs, investigating the previously attempted downstream tasks as well as other tasks, e.g., MRC.
We design LPA to meet various interaction needs of the downstream tasks and propose MSP to avoid the leakages.

In the field of Chinese PLMs, some efforts incorporate coarse-grained information with the character-level inputs.
ERNIE 1.0 \cite{ernie1.0} and BERT-wwm \cite{bert-wwm} propose to mask the words, entities, and phrases as a whole in the MLM task to encourage the modeling of coarse-grained features.
ZEN \cite{zen} adopt auxiliary networks to integrate n-gram representations. 
BERT-MWA \cite{bertmwa} propose word-aligned attention to use multiple segmentation boundaries.
Different from their methods, we propose Lattice-BERT to consume multi-granularity tokens simultaneously into one PLM via lattice graphs.
Thus, Lattice-BERT explicitly exploits the representations of the coarse-grained units, as well as the interactions among word- and character-level tokens.
The proposed MSP task can be treated as an extension of the whole word masking \cite{bert-wwm}, while considering the span information like \citet[SpanBERT]{spanbert} according the lattice structures.
The concurrent work, \citet[AMBERT]{zhang2020ambert} investigate multi-granularity inputs similarly, but they use two transformer encoders to separately deal the word and character sequences.
We treat the words and characters as lattice graphs, which enables thorough interactions among multi-granularity tokens and utilizes all potential segmentation results.

\section{Conclusions}
In this paper, we propose Lattice-BERT to leverage multi-granularity representations of input sentences for Chinese PLMs.
Specifically, Lattice-BERT takes a word-lattice as input, modeling the representations of words and characters simultaneously.
We design the \textit{lattice position attention} to embed the multi-granularity structure into transformers and propose the \textit{masked segment prediction} task to avoid potential leakage in original MLM caused by the redundancy information in lattices.
We conduct extensive experiments on 11 Chinese NLU tasks and observe consistent gains over character-level baselines, achieving new \textit{sota} on CLUE benchmarks.
We show that Lattice-BERT can well manage the lattice inputs and utilize multi-granularity representations to augment the character-level inputs.
We believe the lattice structure can be adapted to integrate the phrase and word representations into the word-piece based PLMs in other languages, which we leave for future exploration.

\section*{Acknowledgments}
This work is supported in part by the National Hi-Tech RD Program of China (No.
2020AAA0106600), the NSFC under grant agreements (61672057, 61672058). For any correspondence, please contact Yansong Feng.

\newpage

\bibliography{anthology}
\bibliographystyle{acl_natbib}

\appendix

\newpage

\section{Ethical Considerations}
\label{app:eth}

Similar to other pre-trained language models \cite{strubell-etal-2019-energy}, 
the study of Lattice-BERT inevitably involves lots of computing time/power. 
The incorporation of multi-granularity tokens generally introduces more computational costs than character-level PLMs \cite{zen,zhang2020ambert}. 
In this section, we elaborate our efforts in reducing the energy costs as well as
discuss the energy comparison with the previous state-of-the-art multi-granularity Chinese PLM, \citet[AMBERT]{zhang2020ambert}.

\paragraph{Efforts in Reducing the Energy Costs. }
In this work, our efforts in reducing the energy costs can be summarized in two folds, i.e., (1) adopting more efficient experiment procedures, 
and (2) reporting performances of \textit{lite}-size models to encourage the followers to compare in lightweight architectures.

From the perspective of experiment procedures, we adopt the mixed-precision arithmetic methods to speed up pre-training. 
We also utilize the two-phase pre-training procedures \cite[BERT]{bert}, where the model processes smaller sequence lengths in 90\% steps (See Appendix~\ref{app:imp:pre}).
In downstream tasks, we search the learning rates with relatively larger strides with heuristic strategies to avoid pointless attempts (See Appendix~\ref{app:imp:fine}).
We adopt ablation studies in \textit{lite}-size. 
The lightweight architecture with six-layer models and 128-character inputs could save much power comparing with training the full-length \textit{base}-size models repeatedly. 

From the perspective of reporting strategies, we report the performance of \textit{base}-size models together with \textit{lite}-size models.
As far as we know, all previous Chinese PLMs only report \textit{base}- or \textit{large}-size settings \cite{wei2019nezha,zen,ernie2,cui2020revisiting,zhang2020ambert}.
Thus, the followers have to implement at least a 12-layer pre-training model to make a fair comparison.
This reporting of \textit{lite}-size performances facilitate the followers to fast validate in \textit{lite}-settings by directly comparing with our LBERT\textit{-lite} and BERT-our\textit{-lite}.
For the text classification and sequence labeling tasks, we use the model trained with 128-sequence length. Thus the followers could make a fair comparison with our methods without pre-training a full-length model (See Appendix~\ref{app:imp:fine}). 
Specifically, the 128-length LBERT-lite model (33M parameters) only costs 3.8 days pre-training with 8 $\times$ NVIDIA V100 16G cards, 1/4 of the original \textit{base-}size Google-BERT model.\footnote{We estimate the pre-training cost of Google-BERT according to Table-3 in \citet{strubell-etal-2019-energy}.}
In conclusion, by reporting the performances of \textit{lite}-size models, we encourage followers to compare with our models with less computational costs.

\paragraph{Energy Comparison.}
We make the energy comparison between LBERT and the previous state-of-the-art multi-granularity Chinese PLM, \citet[AMBERT]{zhang2020ambert}, from the following perspectives: (1) How many computational costs are introduced to process the additional coarse-grained word-level inputs in pre-training. (2) How many additional parameters are introduced by the word-level vocabulary comparing with vanilla fine-grained Chinese PLMs.

Specifically, from the view of additional computational costs,
comparing with the character-level BERT with other settings fixed, 
AMBERT \cite{zhang2020ambert} introduces 100\% more computational costs by adopting two encoders to deal with word- and character-level inputs separately.
Our proposed LBERT-base adopts a uniform architecture to consume words and characters from word-lattices simultaneously. 
Therefore, LBERT only introduces 48\% more computational costs comparing with the corresponding character-level baselines (See Appendix~\ref{app:imp:pre}).
On the other side, from the view of parameter scales, AMBERT has 176M parameters, where additional 68M parameters are introduced by the embedding of word-level tokens.
With the embedding decomposition trick, LBERT only introduces 10M parameters by the word-level units in vocabulary.
As a result, LBERT-base has 100M parameters, comparable to the character-level BERT-our (90M) and RoBERTa (102M) models, and much smaller than AMBERT (176M).
In conclusion, with fewer parameters and smaller computational costs, LBERT is more efficient than AMBERT.

\section{Dataset Statistics}
\label{app:data}

In the \textbf{Tasks} section, we list eleven downstream tasks. 
This note presents several basic statistics, including the sentence length in characters, numbers of tokens in lattices, and dataset scales, which is shown in Table~\ref{Table-tasks}.
For the cloze task, ChID, we focus on a small context of each blank in fine-tuning, thus, we do not report the statistics of the whole passages. 

We show the number of training instances and the training/development/test split in Table~\ref{Table-tasks} as well. 
Specifically, we count the number of questions/blanks in MRC tasks, the number of words/entities for CWS and NER tasks, and the numbers of sentences/sentence pairs in the CLUE classification tasks.
Following the previous effort \cite{zen}, we use the test set for the validation of the MSR-CWS dataset.

\section{Implement Details}
\label{app:imp}
Our code is available at \url{https://github.com/alibaba/AliceMind/tree/main/LatticeBERT}.
Here, we specify some issues in pre-training and fine-tuning.

\subsection{Pre-training Details.}
\label{app:imp:pre}

\paragraph{Lattice-BERT models.}
We train Lattice-BERT with Adam optimizer \cite{kingma2014adam}.
The hyper-parameters of the Lattice-BERT models are shown in Table~\ref{hyperpara}.
The ratio of masked tokens in language modeling is 15\%.
The embedding sizes are different from hidden sizes because we factorize embedding matrices following \citet[ALBERT]{albert} to reduce the additional parameters introduced by word-level inputs. 
For the \textit{base}-size models, the numbers of parameters in the character-level BERT-our and LBERT are 90M and 100M, respectively, less than the corresponding RoBERTa-base (102M) and AMBERT-base (176M).
The numbers of parameters for \textit{lite}-size models are 23M and 33M for BERT-our and LBERT models,  respectively.

\begin{table}[]
\setlength{\tabcolsep}{8pt}
\small
\centering
\begin{tabular}{lcc}
\toprule
Hyper-param&LBERT-base&LBERT-lite\\
\midrule
Number of Layers & 12 & 6 \\
Hidden Size & 768 & 512 \\
Embedding Size & 128 & 128 \\
FFN Inner Hidden Size & 3072 & 2048 \\
Attention Heads & 12 & 8 \\
Attention Head Size & 64 & 64 \\
Dropout & 0.1 & 0.1 \\
Attention Dropout & 0.1 & 0.1 \\
Activation Func. & GELU & GELU \\
Warmup Steps & 5K & 5K \\
Peak Learning Rate & 6e-4 & 6e-4 \\
Batch Size & 8192 & 8192 \\
Max Steps & 100K & 100K \\
Learning Rate Decay & Linear & Linear \\
Adam $\epsilon$ & 1e-6 & 1e-6 \\
Adam $\beta_1$ & 0.9 & 0.9 \\
Adam $\beta_2$ & 0.999 & 0.999 \\
\bottomrule
\end{tabular}
\caption{
Hyper-parameters for pre-training.
\label{hyperpara}}
\end{table}

\begin{table*}[]
\setlength{\tabcolsep}{7.5pt}
\small
\centering
\begin{tabular}{lrrrrrrrrrr}
\toprule
\multirow{2}{*}{Task}   
& \multicolumn{4}{c}{Sentence Length} &
\multicolumn{3}{c}{Dataset Scale}  & \multicolumn{3}{c}{Hyper-parameters}  \\ \cmidrule{2-11} 
&tp995-C&tp995-T&Avg-C&Avg-T&Train&Dev.&Test&max-len.&\#ep.&bs\\ \midrule
\multicolumn{11}{c}{CLUE Text Classification tasks} \\
\midrule
CMNLI   & 136          & 179     & 56.1   & 74.1   & 391.8K&12.2K&13.9K       & 256     & 4     & 16    \\
TNEWS   & 104          & 133     & 41.2   & 53.0   & 53.4K&10.0K&10.0K       & 256     & 5     & 16    \\
IFLY.   & 1028          & 1316     & 291.1   & 384.9   & 12.1K&2.6K&2.6K       & 512     & 5     & 8    \\
AFQMC   & 78          & 106     & 29.7   & 39.6   & 34.3K&4.3K&3.9K       & 256     & 5     & 16    \\
WSC.   & 147          & 187     & 71.7   & 91.8   & 1.2K&0.3K&0.3K       & 256     & 10     & 16    \\
CLS   & 812          & 949     & 298.3   & 384.4   & 20.0K&3.0K&3.0K       & 512     & 5     & 8    \\
\midrule
\multicolumn{11}{c}{CLUE MRC tasks} \\ \midrule
CMRC   &991           & 1301     & 526.7   & 662.9   & 10.1K&1.0K&3.2K       & 512     & 5     & 8    \\
ChID   & --          & --     & --   & --   & 577.2K&23.0K&23.0K       & 64     & 2     & 24    \\
C$^3$   & 1088          & 1479     & 224.9   & 329.1   & 11.9K&4.3K&3.9K       & 512     & 8     & 12    \\ \midrule
\multicolumn{11}{c}{Sequence Labeling tasks} \\ \midrule 
MSR-CWS   & 171          & 238     & 48.6   & 66.6   & 2.37M&-&106.9K       & 512     & 5     & 8    \\
MSRA-NER  & 174          & 247     & 49.0   & 67.1   & 34.0K&3.8K&7.7K     & 512     & 10     & 8     \\ 
\bottomrule
\end{tabular}
\caption{
Statistics of the datasets and hyper-parameters for downstream tasks.
tp995-C and tp995-T are the top 99.5\% length of the input sequence in characters and the top 99.5\% number of lattice tokens, respectively. 
Avg-C and Avg-T are the average numbers of the characters and the lattice tokens per instance. 
max-len., \#ep., and bs represent the max sequence length of the BERT model (max lattice tokens that Lattice-BERT models consume), the number of epochs, and the batch size, respectively.
\label{Table-tasks}}
\end{table*}

To speed up the pre-training, we adopt the two-phase procedures of \citet[BERT]{bert}, which first pre-trains the model with a sequence length of 128 characters per instance for 90\% steps, and then trains the rest 10\% steps with a sequence length of 512 characters per instance.
We base our code on the optimized version of BERT released by NVIDIA by leveraging the mixed-precision arithmetic and the multi-GPU techniques.
For the base-size LBERT models, the two pre-training phases take about 9.2 and 6.7 days, separately, with NVIDIA 8 $\times$ V100 16G cards.
For lite-size models, the time consumption for the two phases take 4.0 and 3.1 days. 
We discuss the energy resources in the Ethical Considerations section.

To make fair comparison, we expand the maximum size of input tokens in pre-training of LBERT to process the additional word-level lattice tokens, following previous multi-granularity PLMs in Chinese \cite{zen,zhang2020ambert}.\footnote{Under the context of Lattice-BERT, the \textit{input token length} or the \textit{max sequence length} means the maximum number of lattice tokens per instance that the Lattice-BERT consumes.}
Particularly, we ensure each training instance of character-level baselines (i.e., BERT-our) and the Lattice-BERT models contains the same number of character-level tokens, which is 128 and 512 for the first and second pre-training phase, respectively.
As a result, from the view of the pre-training data creation, fixing the corpus size, the numbers of instances per epoch are the same between character-level BERT-our and LBERT in pre-training given the same corpus. 
Thus, in the same steps, we can train BERT-our and LBERT for the same number of epochs.
In another word, BERT-our and LBERT process the same size of corpus with the same training steps.

In practice, based on the statistics on the pre-training corpus, we expand the input size by 35\%. 
For example, the instances in the first pre-training phase of LBERT have 173 lattice tokens, where 128 Chinese characters are expected to be contained.
For the second pre-training phase of LBERT, the input token size is 692.

However, via an empirical estimation, this extension makes the pre-training procedure of the Lattice-BERT models cost 48\% more computational resources than the corresponding BERT-our settings.\footnote{Theoretically, the time complexity of the attention layers to the input length is $O\left(n^2\right)$, and that of the fully connected layers is $O\left(n\right)$. Thus, the overall increment of time cost is between 35\% and 82\% (1.35$\times$1.35$\approx$1.82). To empirically estimate the complexity, we use the ratio of pre-training time between taking 128 and 173 input lengths separately, under the setting of \textit{lite}-size BERT-our.}
This increment of time complexity is much lower than other multi-granularity PLMs like AMBERT \cite{zhang2020ambert}, which introduces 100\% additional computational resources comparing with the corresponding character-level setting. 
We further compare the performances of BERT-our and LBERT in the downstream tasks under the same pre-training computational costs in Appendix~\ref{app:exp}, where LBERT still outperforms BERT-our by a large margin.

\paragraph{BERT-our Baselines.}
The pre-training settings of BERT-our are almost the same as our proposed LBERT.
These settings provide a fair comparison to support the argument that the improvements are attributed to better utilizing the multi-granularity information in word lattices.
The differences between BERT-our and LBERT in model architectures are: 
(1) BERT-our use the same vocabulary and tokenization functions as the Chinese version of Google BERT \cite{bert}, and taking character level inputs, while LBERT takes word lattices as inputs. 
(2) BERT-our has shorter input sequences to ensure the Lattice-BERT and BERT-our
training for the same epochs in the same steps.
(3) LBERT incorporates the positional relations between lattice tokens in attention layers, which is not suitable for BERT-our with sequential inputs. 
The other components in \textit{lattice position attention}, including the absolute position attention and the distance information, are adopted in BERT-our as well.
(4) LBERT adopts \textit{whole segment prediction} to avoid potential leakage, while BERT-our adopts the whole word masking \cite[wwm]{bert-wwm} trick to utilize the word level information.

\begin{table*}[]
\setlength{\tabcolsep}{6.5pt}
\small
\centering
\begin{tabular}{l|ccc|ccc|ccc|ccc}
\toprule
dataset
&best lr.&avg.&std.
&best lr.&avg.&std. 
&best lr.&avg.&std. 
&best lr.&avg.&std. 
\\ \midrule
&\multicolumn{2}{l}{BERT-our-\textit{base}}&&\multicolumn{2}{l}{BERT-our-\textit{lite}}&&\multicolumn{2}{l}{LBERT-\textit{base}}&&\multicolumn{2}{l}{LBERT-\textit{lite}}\\ \midrule
CMNLI & 1.5e-5 & 80.6 & 0.4 & 8.0e-5 & 77.6 & 0.2 & 1.0e-5 & 81.6 & 0.2 & 3.0e-5 & 79.2 & 0.2 \\
TNEWS & 1.5e-5 & 66.7 & 0.2 & 5.0e-5 & 65.4 & 0.2 & 3.0e-5 & 67.8 & 0.3 & 3.0e-5 & 67.2 & 0.2 \\
IFLY. & 3.0e-5 & 60.8 & 0.9 & 1.0e-4 & 59.3 & 0.2 & 2.0e-5 & 61.7 & 0.1 & 8.0e-5 & 60.0 & 0.3 \\
AFQMC & 2.0e-5 & 74.3 & 0.5 & 5.0e-5 & 72.3 & 0.6 & 1.0e-5 & 74.6 & 0.3 & 3.0e-5 & 73.1 & 0.4 \\
WSC. & 5.0e-5 & 81.5 & 1.3 & 1.5e-4 & 66.3 & 1.9 & 2.0e-5 & 85.5 & 1.1 & 8.0e-5 & 75.3 & 1.3 \\
CSL & 1.5e-5 & 81.4 & 0.3 & 1.0e-4 & 78.5 & 0.7 & 1.0e-5 & 83.3 & 0.2 & 1.0e-4 & 81.4 & 0.4 \\
\midrule
CMRC/F1 & 5.0e-5 & 86.9 & 0.2 & 1.5e-4 & 83.7 & 0.4 & 3.0e-5 & 87.4 & 0.5 & 8.0e-5 & 84.9 & 0.2 \\
CMRC/EM & - & 67.6 & 0.4 & - & 62.6 & 0.4 & - & 68.3 & 0.6 & - & 64.9 & 0.6 \\
ChID & 1.5e-5 & 84.6 & - & 5.0e-5 & 79.0 & - & 1.0e-5 & 86.4 & - & 2.0e-5 & 81.7 & - \\
C$^3$ & 5.0e-5 & 69.1 & 0.3 & 5.0e-5 & 61.9 & 0.6 & 5.0e-5 & 72.7 & 0.4 & 8.0e-5 & 63.0 & 0.5 \\
\midrule
MSRA-NER/EF1 & 2.0e-5 & 94.6 & 0.8 & 5.0e-5 & 92.8 & 0.2 & 5.0e-5 & 95.6 & 0.2 & 5.0e-5 & 94.1 & 0.1 \\
MSRA-NER/LF1 & - & 96.5 & 0.5 & - & 95.5 & 0.1 & - & 97.1 & 0.1 & - & 96.2 & 0.1 \\
MSR-CWS & 3.0e-5 & 98.4 & 0.0 & 8.0e-5 & 98.1 & 0.0 & 8.0e-5 & 98.6 & 0.0 & 8.0e-5 & 98.4 & 0.0 \\
\midrule
Average & - & 79.1 & - & - & 75.1 & - & - & 80.6 & - & - & 77.2 & - \\
\bottomrule
\end{tabular}
\caption{ 
The full experimental results, with best learning rates (best lr.), corresponding average (avg.) and standard deviation (std.) for five runs. For the NER task, EF1 and LF1 are entity-level F1 scores and label-level F1 scores, respectively. We average the multiple metrics within each task (e.g., CMRC) before averaging over tasks.
\label{allresults}
}
\end{table*}

\paragraph{Vocabulary. }
The vocabulary used to construct lattices is a superset of the vocabulary in BERT-our and the Chinese version of Google-BERT \cite{bert}.
Particular, the vocabulary of LBERT consists of 21K tokens (including characters and word pieces) from the vanilla vocabulary of BERT-our and 81K additional high-frequency words from the pre-training corpus. 
To obtain the high-frequency words, we randomly sample 10\% of the pre-training corpus, running an in-house built tokenizer, and counting the word frequency after tokenization.
All the English tokens in this vocabulary are lower-cased, which means LBERT can be seen as an \textit{uncased} model.
However, LBERT is a Chinese PLM, thus, only a few English tokens exist in pre-training and fine-tuning.

\subsection{Fine-tuning Details.}
\label{app:imp:fine}

\paragraph{Hyper-Parameters and Settings}
We tune the learning rates in (8e-6, 1e-5, 1.5e-5, 2e-5, 3e-5, 5e-5, 8e-5, 1e-4, 1.5e-4) on the development sets with the other hyper-parameters, including max sequence length of the BERT model (max-len), the number of epochs (ep.), and the batch size (batch), fixed. 
The hyper-parameters in fine-tuning are shown in Table~\ref{Table-tasks}. 

Specifically, for the sequence labeling tasks, including CWS and NER, we choose the best learning rates based on the label-level F1 scores on the development sets.
In practice, we adopt heuristic strategies to avoid pointless attempts. 
For example, if a model performs monotonic increasing in a down-stream task with the learning rates of 2e-5, 3e-5, and 5e-5, we will not try the learning rates lower than 2e-5.
The best learning rates, together with the average and the standard deviation of the scores for five runs, are shown in Table~\ref{allresults}. 
We report the average and deviation scores on development sets for CLUE tasks and those on the test sets for sequence labeling tasks (NER \& CWS).
In the ChID task, we run the models once with the best learning rates for efficiency, because its training set is too large.

In text classification and sequence labeling tasks, the max sequence length are the same between Lattice-BERT and the character-level BERT-our, thus, their time complexities are the same in fine-tuning.
However, in MRC tasks, where the documents are relatively longer, it is essential to read the long passages as complete as possible. 
Thus, to make fair comparisons, we expand the size of input lattices in LBERT settings to ensure the numbers of Chinese characters per instances are the same between LBERT and BERT-our settings.

\begin{table*}[t]
\small
\centering
\setlength{\tabcolsep}{4.8pt}
\begin{tabular}{l rrr | cccccc|c|cc|c}
\toprule
 &&&&
\multicolumn{7}{c|}{CLUE-Classification} & \multicolumn{2}{c|}{Seq. Lab.} & \multirow{3}{*}{\shortstack{\\avg.}}  \\ \cmidrule{5-13} 
 &&&&
  NLI &
  \multicolumn{2}{c}{TC} &
  \multicolumn{1}{c}{SPM} &
  \multicolumn{1}{c}{CoRE} &
  \multicolumn{1}{c|}{KwRE} &
  \multirow{2}{*}{ \shortstack{\scriptsize avg.}} &
  CWS &
  \multicolumn{1}{c|}{NER} &
  \\ 
  \cmidrule{5-10}
  \cmidrule{12-13}
 & 
 \scriptsize Para. & \scriptsize Cmplx. & \scriptsize Epoch. & 
  \multicolumn{1}{c}{\scriptsize CMNLI} &
  \scriptsize TNEWS &
  \multicolumn{1}{c}{\scriptsize IFLY.} &
  \scriptsize AFQMC &
  \multicolumn{1}{c}{\scriptsize WSC.} &
  \multicolumn{1}{c|}{\scriptsize CSL} & &
  MSR &
  \multicolumn{1}{c|}{\scriptsize MSRA} \\
\midrule 
 BERT-our & 23M & T & N &
  77.6 &
  65.4 &
  59.3 &
  72.3 &
  66.3 &
  78.5 &
  69.9 &
  98.1 &
  95.5 &
  76.7  \\ 
\quad \scriptsize +seql & 23M & 1.48T & 1.35N &
  77.6 &
  65.3 &
  59.3 &
  72.4 &
  68.4 &
  80.1 &
  70.5 &
  98.1 &
  95.6 &
  77.1  \\ 
\quad \scriptsize +seql-EmbDe & 31M & 1.48T & 1.35N &
  78.4 &
  65.8 &
  59.7 &
  71.6 &
  71.4 &
  79.9 &
  71.1 &
  98.3 &
  95.8 &
  77.6  \\ \midrule
 LBERT & 33M & 1.48T & N &
  \textbf{79.2} &
  \textbf{67.2} &
  \textbf{60.0} &
  \textbf{73.1} &
  \textbf{75.3} &
  \textbf{81.4} &
  \textbf{72.7} &
  \textbf{98.4} &
  \textbf{96.2} &
  \textbf{78.9}  \\ 
\bottomrule
\end{tabular}
\caption{\label{Table-app:exp}
The performances on development sets in \textit{lite}-size settings. 
Para., Cmplx., and Epoch. are the parameter sizes, time complexities, and the training epochs, respectively. We mark the time complexity and the training epochs of the BERT-our as T and N.
}
\end{table*}

\paragraph{Selection of Pre-training Models.}
We adopt the PLMs after the first pre-training phase, i.e., pre-training with the 128-character input size, for the text classification, sequence labeling, and ChID tasks.
The details of the two-phase pre-training procedure refer to Appendix~\ref{app:imp:pre}.
Pilot experiments demonstrate that, for these tasks, PLMs with shorter pre-training lengths perform comparably with the full-length versions.
We think this is because the sentence lengths are relatively small in these tasks except for IFLY., the long text classification task, where the first few sentences are more crucial for the predictions. 
This strategy makes our explorations much more efficient. 
We adopt the full-length pre-training versions for the CMRC and C$^3$ tasks.

\paragraph{Detailed Implementation.}
For the implementation of fine-tuning tasks, we adopt the simplest methods following the CLUE official examples.\footnote{\url{https://github.com/CLUEbenchmark/CLUE/tree/master/baselines/models_pytorch}}
For example, we adopt logistic regressions over the \texttt{[CLS]} tokens for classifications and softmax regressions over the \texttt{[CLS]} tokens of all options for multiple choices in C$^3$ and ChID. 
To deal with the long documents in MRC tasks, we truncate document to at most 512 characters in C$^3$.
For CMRC, we split the document into segments of at most 512 characters with the stride of 128 characters. 
Then the selected spans from each segment are aggregated to find the final answer to the question.
Furthermore, for ChID, the contexts no far than 32 characters to the masked idiom are incorporated.

We conduct data augmentation for TNEWS and CSL tasks.
In TNEWS, we use both keywords and titles for classifications. Moreover, for the CSL task, we concatenate all the keywords.
These augmentations are also conducted in the previous works, which is either explicitly mentioned in the paper \cite[AMBERT]{zhang2020ambert}, or can be inferred from the performances \cite[NEZHA]{wei2019nezha}.\footnote{The TNEWS performance in NEZHA is 67.4\%. 
However, without data augmentation, even the large models like ALBERT-xxlarge and RoBERTa-wwm-large could not obtain an accuracy more than 60\% \cite{CLUEbenchmark}.}

\section{Further Experiments}
\label{app:exp}

Mentioned in \S3.4 and Appendix~\ref{app:imp:pre},  we expand the maximum size of input tokens in pre-training of LBERT to process the additional word-level lattice tokens, following previous multi-granularity PLMs in Chinese \cite{zen,zhang2020ambert}.
We expand the input token length for the pre-training of LBERT by 35\%, which makes the LBERT and BERT-our have the same training epochs when the training steps are equal, but introducing 48\% more computational resources (discussed in Appendix~\ref{app:imp}).
Meanwhile, the additional word-level tokens in the vocabulary of LBERT introduce 11\% and 43\% more parameters in the embedding matrix for the \textit{base}-size and \textit{lite}-size settings, respectively.
To illustrate the gains of LBERT attribute to the incorporation of lattices instead of additional computations or parameters, we investigate the BERT-our with longer input sequences in pre-training and the BERT-our model without the embedding decomposition trick, which has more parameters in embedding.

To reduce computational costs, we base the experiments on the first pre-training phase and the \textit{lite}-size setting (see Appendix~\ref{app:imp:pre} for details). 
Since we do not conduct full-length pre-training, we use CLUE classification and sequence labeling tasks. 
We report the \textbf{\textit{average scores}} on development sets over five runs. 
The compared system is listed below: 

\noindent\textbf{+seql} is the character-level BERT-our taking the same sequence length (i.e., 173) as LBERT. Comparing with LBERT, this setting results in the same time consumption, together with 35\% more character-level tokens per instance comparing to LBERT models, which results in 35\% more corpus/epochs to process with the same training steps.

\noindent\textbf{-EmbDe} is the character-level BERT-our without embedding decomposition tricks, i.e., the size of embedding matrix is the same as the hidden size, 512. 
In the \textit{lite}-size setting, wihtout embedding decomposition, the character-level BERT-our have 8M more parameters, and the total parameter scale is 31M, comparable to that of LBERT (33M).

\paragraph{Results.} As we can see in Table~\ref{Table-app:exp}, LBERT remarkably outperforms the BERT-our settings, even after the extension of time complexity and embedding parameters.
Specifically, on CLUE classification tasks, LBERT outperforms BERT-our+seql
and BERT-our+seql-EmbDe by 2.2\% and 1.6\%, respectively. 
On sequence labeling tasks, the improvements of +seql and -EmbDe are also marginal.
These results demonstrate that the performance gains of LBERT attribute to the usage of multi-granularity representations in word lattices instead of the additional pre-training time complexities and embedding parameters.

\end{CJK}
\end{document}